\documentclass[conference]{IEEEtran}
\usepackage{multirow} 
\usepackage{multicol}
\IEEEoverridecommandlockouts
\usepackage{cite}
\usepackage{amsmath,amssymb,amsfonts}
\usepackage{algorithmic}
\usepackage{graphicx}
\usepackage{textcomp}
\usepackage{xcolor}
\usepackage{float}
\usepackage{hyperref}

\def\BibTeX{{\rm B\kern-.05em{\sc i\kern-.025em b}\kern-.08em
    T\kern-.1667em\lower.7ex\hbox{E}\kern-.125emX}}
\usepackage{xspace}
\usepackage{lipsum}
\usepackage{subfigure} 

\usepackage{color, xcolor}
\usepackage[linesnumbered,ruled]{algorithm2e}
\usepackage{booktabs}
\usepackage{multirow}
\usepackage{threeparttable}
\usepackage{array}
\usepackage{bbding}
\usepackage{svg}
\usepackage{epigraph} 
\usepackage{balance}

\begin{document}
\newcommand{\sys}{{\sc Protego}\xspace}
\title{{\sys}: Detecting Adversarial Examples for Vision Transformers via Intrinsic Capabilities\\

}

\author{\IEEEauthorblockN{Jialin Wu,
Kaikai Pan,
Yanjiao Chen\IEEEauthorrefmark{1}, 
Jiangyi Deng,
Shengyuan Pang, and
Wenyuan Xu}
\IEEEauthorblockA{USSLAB, Zhejiang University \\
Email: \{jialinwu, pankaikai, chenyanjiao, jydeng, pangpang0093, wyxu\}@zju.edu.cn}

\thanks{\IEEEauthorrefmark{1}Corresponding author: Yanjiao Chen, chenyanjiao@zju.edu.cn.}
}




\maketitle

\begin{abstract}
Transformer models have excelled in natural language tasks, prompting the vision community to explore their implementation in computer vision problems. However, these models are still influenced by adversarial examples. In this paper, we investigate the attack capabilities of six common adversarial attacks on three pre-trained ViT models to reveal the vulnerability of ViT models. To understand and analyse the bias in neural network decisions when the input is adversarial, we use two visualisation techniques that are attention rollout and grad attention rollout. To prevent ViT models from adversarial attack, we propose \sys, a detection framework that leverages the transformer intrinsic capabilities to detection adversarial examples of ViT models. Nonetheless, this is challenging due to a diversity of attack strategies that may be adopted by adversaries. Inspired by the attention mechanism, we know that the token of model's prediction contains all the information from the input sample. Additionally, the attention region for adversarial examples differs from that of normal examples. Given these points, we can train a detector that achieves superior performance than existing detection methods to identify adversarial examples. Our experiments have demonstrated the high effectiveness of our detection method. For these six adversarial attack methods, our detector’s AUC scores all exceed 0.95. \sys may advance investigations in metaverse security.
\end{abstract}


\section{Introduction}
The metaverse refers to a virtual, expanded, integrated shared space that encompasses multiple user-generated virtual worlds or spaces\cite{b1,b2}. The metaverse is an integral component of the digital economy and represents an integrated innovation of next-generation information technology. With the advent of the intelligent industrial revolution triggered by LLMs, the metaverse is poised for vigorous development due to the capability of LLMs to achieve cross-disciplinary, multilingual, and multimedia intelligence. Furthermore, the metaverse is expected to provide humanity with expanded creative space and content experiences. The vision transformer serves as the foundation for Multimodal Large Language Models and is an indispensable component in the visual domain of the metaverse\cite{b3}. Therefore, the exploration of its security issues is of paramount importance.

\begin{figure}[!t]
\centerline{\includegraphics[width=0.5\textwidth]{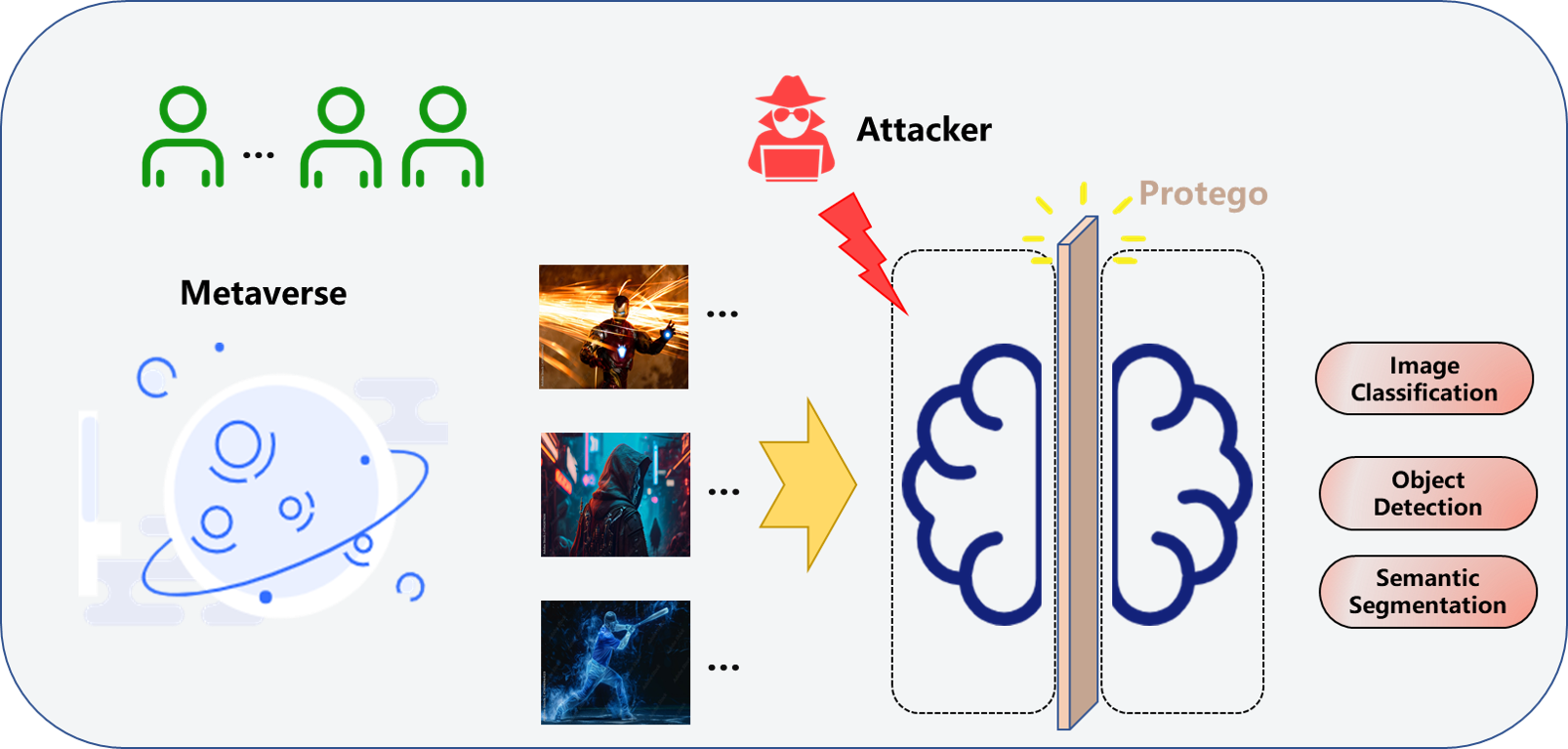}}
\caption{Security issues in the computer vision domain within the metaverse. The Protego is a charm(\textbf{our detector}) that protected the caster(\textbf{model}) with an invisible shield that reflected spells and blocked physical entities(\textbf{adversarial examples}).}
\label{fig_metaverse}
\end{figure} 

The Transformer is an attention-based encoder-decoder architecture that has fundamentally revolutionized the field of natural language processing\cite{b4}. In recent years, the Transformer has garnered significant success across various machine learning tasks, particularly within multimodal applications. The initial inspiration for artificial intelligence is to replicate human perception. With the remarkable performance exhibited by LLMs, the goal of constructing a "human-like brain" model has drawn closer, sparking widespread discussions on multimodal learning. The Transformer architecture represents a competitive family of models, presenting new challenges and opportunities for multimodal machine learning. Notably, the recent success of LLMs and their cross-modal derivative models(\textit{e.g.}, DiT) further underscores the potential of the Transformer in laying the groundwork for Multimodal Large Language Models(MLLMs)\cite{b5,b8}. The emergence of the Vision Transformer is pivotal in transitioning the Transformer from the field of natural language processing to computer vision, serving as a cornerstone for multimodal machine learning.

 Vision Transformer applies the Transformer's encoder to images by dividing the entire image into smaller patches\cite{b6}, which are then linearly embedded into sequences and fed into the Transformer network. This approach enables the training of an end-to-end model for various image-related tasks. Due to the convolution operation in CNN, it can only capture local information and cannot establish long-distance connections for the entire image. Vision Transformer and its variants have been widely utilized across various computer vision tasks, including recognition, detection, and segmentation. However, similar to traditional CNN models, ViTs are also susceptible to adversarial attacks from adversarial examples\cite{b7}.

 In the current research, there has been an extensive exploration of adversarial examples targeting CNN models\cite{b25,b26,b27,b28}. Regarding strategies to counter adversarial examples, two types of defense mechanisms have emerged: (1) Reactive approaches involve detecting adversarial examples and implementing preemptive defense measures. (2) Proactive strategies focus on enhancing model robustness to withstand adversarial attacks. Because of the significant structural differences between CNNs and ViTs, methods that perform well in CNNs are challenging to transfer directly to ViTs. Currently, the focus of adversarial attacks on ViTs is more centered on the transferability of adversarial examples between ViTs and CNNs\cite{b9}.

\begin{figure}[!t]
\centerline{\includegraphics[width=0.35\textwidth]{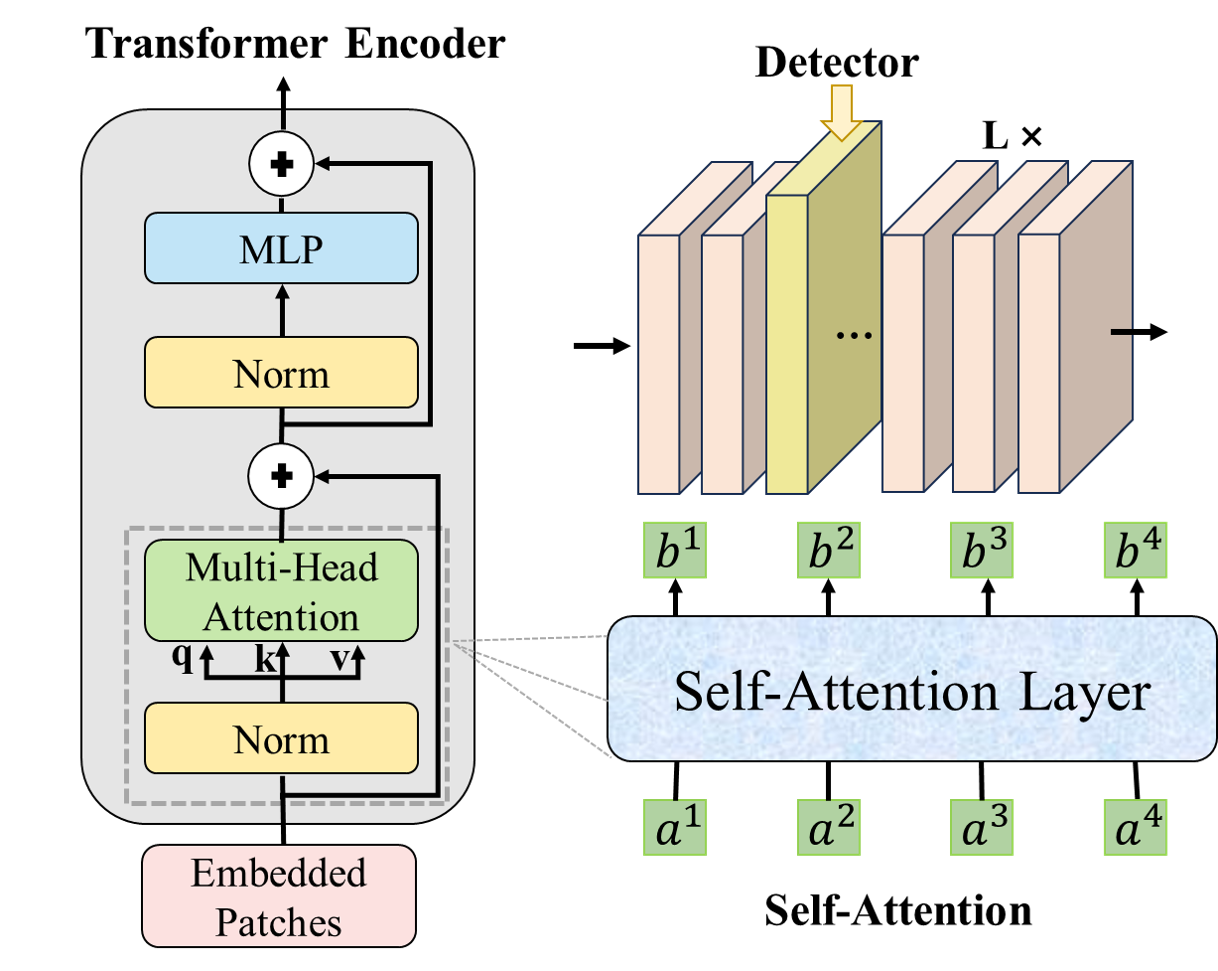}}
\caption{The architecture of Vision Transformer, and the attention mechanism in transformer encoder.}
\label{fig1}
\end{figure}

 The basis of ViT is the self-attention mechanism, with its core design principle centered around recalibrating the input by computing the interrelations among tokens to yield the output\cite{b3,b10}. The key components of ViT are illustrated in Figure.~\ref{fig1}. Within ViT, the input image is initially partitioned into a sequence of non-overlapping patches, each of which is then embedded. To retain spatial information, a one-dimensional learnable positional encoding is incorporated into these patch embeddings. The combined embeddings are subsequently fed into the encoder, with the learned [\textit{cls}] token fused with the patch embeddings to consolidate the global representation, serving as the input for classification. 

 In this paper, we propose a novel detection framework named self-guard, which features powerful. To realize this design goal, we have to investigate two research problems.

\begin{itemize}
    \item \textit{How to design a framework of adversarial example detection that is applicable to ViTs?}
\end{itemize}

After utilizing the structural information of the Vision Transformer (ViT) to perform patch embeddings on images, and passing them through multiple layers of transformer encoders, the feature output of the last layer or a specific layer is obtained. This is used to seek inconsistencies in features between adversarial and normal examples in order to train a lightweight, plug-in detector.

\begin{itemize}
    \item \textit{How to obtain a more effective adversarial sample detector?}
\end{itemize}

As a result of the transformer’s inherent characteristics, an interaction mechanism is established among the network’s various modules, allowing for the fusion of feature maps at different levels across the encoder, decoder, and self-attention fusion module. This enhances the ability to effectively capture the feature information of input, thus distinguishing between adversarial and normal examples.

We apply the framework named \sys of plugin-detector on three different pre-trained models, evaluated with ImageNet, a representative dataset. The comparison with one baselines demonstrates that \sys achieves the best adversarial detection performance.

We summarized the following three contributions:

\begin{itemize}
    \item We propose an adversarial sample detection framework for the ViT model, achieving high-accuracy detection of adversarial examples.
\end{itemize}

\begin{itemize}
    \item We develop a new feature extraction technique in \sys, which is universal for transformer structures.
\end{itemize}

\begin{itemize}
    \item We conduct experiments, demonstrating that \sys outperforms one baseline, and our detector is plug-in.
\end{itemize}

The rest of the paper is organized as follows: Section II introduces the background and related work. Section III presents the technical details of the proposed method. Section IV provides a series of experiments to demonstrate the effectiveness of the proposed method. Section V discusses the limitations of the method and future work, and finally, in Section VI, we conclude the paper.

\section{Background \& Related works}

\subsection{Vision Transformer}

ViT\cite{b6} is the first work to apply the architecture of cascading multiple Transformer layers to mid-sized image datasets such as ImageNet\cite{imagenet}, replacing standard convolutions. They made minor adjustments to the base Transformer model and applied it to a sequence of image “patches” flattened as vectors. A learnable embedding is applied to the sequence of embedding patches, serving the same purpose as the [\textit{class}] token in BERT. This embedding method is used as the image representation. Notably, ViT consists solely of the transformer’s encoder, which is similar to BERT. The encoder-only architecture excels at analyzing and classifying input content. This is also why many studies utilize ViT as a model for image classification and recognition. In general, due to computational resource constraints, many ViT models are fine-tuned for specific downstream tasks using pre-trained models on large datasets like ImageNet-21K\cite{b24}.

Touvron \textit{et al.} \cite{b11}proposed competitive convolution-free transformers(DeiT) trained on ImageNet with few overheads. The vision transformer has the same architecture as ViT-B and employs 86M parameters. DeiT is a fully Transformer-based architecture. Its core is the proposal of a teacher-student distillation training strategy for ViT, and the introduction of token-based distillation methods, enabling Transformers to be trained quickly and effectively in the visual domain. This model has performed very well on ImageNet. It achieves top-1 accuracy of 83.1\%(single-crop) on ImageNet with no external data. Swin Transformer is a variant based on the ViT model\cite{b12}. By introducing a sliding window mechanism, the model can learn information across windows. Additionally, through downsampling layers, the model can process high-resolution images, thus saving computational resources and focusing on global and local information. A novel Transformer-in-Transformer (TNT) architecture has been proposed for visual recognition\cite{b13}, focusing on local information within patches. It goes beyond simply dividing the image into a series of patches, instead opting for a more granular approach by reshaping each patch into a sequence of pixels. Twins have improved the attention mechanism by integrating local-global attention\cite{b14}, grouping the spatial dimensions of features to compute self-attention within each group, and then merging the attention results from the global perspective across the grouped attention.

In this paper, we choose three ViT models, two ViT models are general models, and one is variant of original model.

\subsection{Adversarial attack}

For a DNN, intentionally adding imperceptible perturbations to input can lead the model to confidently produce an incorrect output\cite{b15}. Adversarial attacks can be categorized as white-box and black-box.

\paragraph{White-Box Attacks}

White-box attacks require full knowledge of the target model. Although not practical in real-world scenarios, they better expose the vulnerability of deep learning models. The Fast Gradient Sign Method (FGSM) is one of the most well-known untargeted attack methods and can be directly extended to targeted attacks\cite{b15}. It is a single-step optimization method that involves gradient ascent, with the optimization direction being opposite to the gradient descent of the trained model. BIM\cite{b16}, which stands for Basic Iterative Method, adds iterative operations on top of FGSM. The Basic Iterative Method (BIM) attack is an iterative attack and can be seen as a variant of the FGSM. FGSM involves a single iteration (single step), while BIM involves multiple iterations (multiple steps), taking a small step each time. During each iteration, the perturbation is adjusted within a specified range. PGD can also be viewed as multiple iterations of FGSM\cite{b17}. However, the difference between PGD attack and BIM is that PGD initializes with uniform random noise, which adds an initial random perturbation compared to BIM.

\paragraph{Black-Box Attacks}

Black-box attacks need less knowledge than white-box attacks. Dong\textit{et al.} proposed a class of momentum-based iterative algorithms to enhance adversarial attacks\cite{mim}. By integrating a momentum term into the iterative process of the attack, our method can stabilize the update direction and break free from undesirable local maxima during the iterations, thus generating more transferable adversarial examples. A black-box attack method based on surrogate models and the transferability of adversarial sample was proposed by Papernot\textit{et al.}\cite{b19}. They successfully achieved black-box attacks on the MNIST dataset using a training method with Jacobian-based data augmentation. 

We discuss six adversarial attacks in this paper, including both white-box attacks and black-box attacks.

\subsection{Defense}

Defense methods can be divided into improving the robustness of the model and detecting and rejecting adversarial examples to protect the model from attack. 

\paragraph{Detection} Feature squeezing reduces the adversary's search space by combining the examples that correspond to many different feature vectors in the original space into a single sample\cite{fs}. It can accurately detect adversarial examples by comparing the DNN model's predictions on the original input with those on the input after feature squeezing. Ma\textit{et al.} proposed Local Intrinsic Dimensionality (LID) to detect adversarial examples\cite{lid}. When fitting data, neural networks have a tendency to fit in directions that are easier to fit by gradient descent. This makes it easier for neural networks to fit low-dimensional data sets, regardless of the density and distance of the data in a given space. Adversarial examples, on the other hand, often require higher dimensions to explain than normal examples. Since adversarial examples are distributed outside the manifold of normal data, two new features, namely Kernel Density Estimation(KD) and Bayesian Uncertainty Estimation(BU), are proposed to detect adversarial examples\cite{kd}. KD is used to detect and analyze points far from the data manifold by computing points in the feature space of the last hidden layer. BU can be applied to dropout neural networks to monitor when points fall into low-confidence regions of input space, detecting adversarial examples that KD cannot.

Overall, the aforementioned detection methods lack generalizability and are difficult to defend against adaptive attacks. In this paper, our method can effectively withstand adaptive attacks and defend against previously unseen attack methods.

\paragraph{Model enhancement} Adversarial training involves adding adversarial examples to the training set\cite{b32}, and the feature distribution of the adversarial examples can be learned as the model is trained, thereby enhancing the robustness of the model. Adversarial training can be summarized by the following maximization-minimization formula.

\begin{equation}
    \label{eq:min_max}
  \min_{\theta} \mathbb{E}_{(z,y)\sim D} \left[ {\max_{||\delta||\leq\epsilon} L(f_\theta(X+\delta),y)} \right] 
\end{equation}
Where $\delta$ is noise, and $\epsilon$ is bound for perturbation.

Madry \textit{et al.}\cite{b33} proposed PGD adversarial training to make models that are robust to adversarial inputs. Kannan \textit{et al.} \cite{b34} proposed the mixed-minibatch PGD (M-PGD) adversarial training method, which incorporates a logit pairing strategy based on PGD adversarial training \cite{b33}. M-PGD incorporates clean examples in the adversarial training, and the logit pairing strategy involves the pairing of a normal example with an adversarial example or another normal example. However, Adversarial training will compromise the performance of the original model.

\section{Design}

\begin{figure*}[htbp]
\centerline{\includegraphics[width=0.67\textwidth]{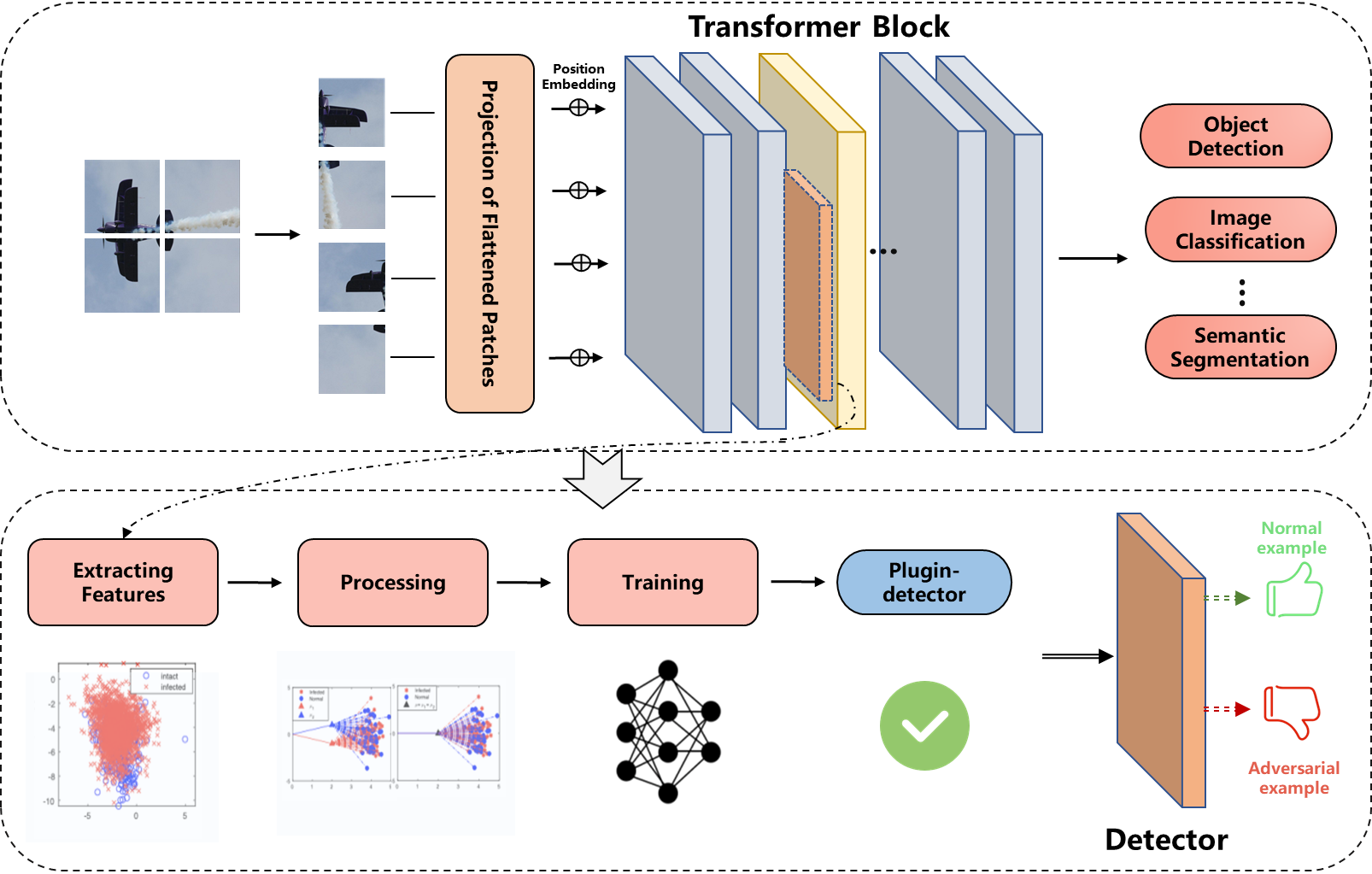}}
\caption{Framework of \sys. The detector is trained on the features in the transformer and can be inserted into the layer that extracts the features of the transformer block.}
\label{fig2}
\end{figure*}

In this work, we propose a framework named \sys to detect adversarial examples for vision transformers. As illustrated in Figure.~\ref{fig2}, the upper part shows the workflow of ViT, where the Transformer consists only of the encoder part, and it is composed of multiple layers cascaded together. The lower part presents the technical solution of \sys.

\subsection{Features Extracting}\label{AA}

Because of the ability of the transformer encoder, normal examples and adversarial examples, when passed through the transformer layer as input, will exhibit different behaviors in high-dimensional features. Transformer is a Sequence to Sequence model, notable for its extensive use of self-attention. As shown in Figure~\ref{fig1}, self-attention is a new type of neural network layer. Its input and output are identical to those of RNN: it takes in a sequence and produces a sequence. For each output, $b^{i}$ ($i=1,2,3,4$), it has considered the entire input, denoted as $a^{i}$ ($i=1,2,3,4$). The pipeline of the transformer can be demonstrated by the following formula.

\begin{align}
  q^i &= W^q a^i \\ 
  k^i &= W^k a^i \\ 
  v^i &= W^v a^i 
\end{align}  
Which $a^i$ is the embedding of input, and $W^q, W^k, W^v$ are three different transformation matrices.

Then, to calculate the attention magnitude, use each query in turn to perform a scaled inner product with each key.

\begin{equation}
    \label{eq:4}
    B = Attention(q,k,v) = softmax(\frac{q k^T}{\sqrt{d_k}})v
\end{equation}
In Equation~(\ref{eq:4}), $\sqrt{d}$ is to transform the attention matrix into a standard normal distribution, and $B$ stands for the output $b^i$($i=1,2,3,4$). 

Therefore, $b^1$ carries all the information of the input's embedding. We extract the matrix $b^1$ from the encoder's output by inputting normal examples and adversarial examples, successively. This matrix serves as the feature distribution required for our training, denoted as $D_{clean}(x)$ and $D_{adv}(x)$. We regard $D_{adv}(x)$ and $D_{clean}(x)$ as the features of examples, training the detector on the train set.

\begin{equation}
    \label{eq:5}
  \textit{features}_{noise} = D_{adv}(x) - D_{clean}(x) 
\end{equation}
$\textit{features}_{noise}$ represents the distinguishing features of the adversarial examples. And we know this is the features of noise, it also works in this paper.

\subsection{Architecture of Plugin-Detector}

In our paper, the architecture of plugin-detector is one-layer linear neural network, a simple yet efficient architecture. 

\begin{equation}
    \label{eq:6}
  \textit{Detector}(f) = \text{Linear}\left(\text{reshape}(f,[1, m \times n])\right)
\end{equation}
Where f represents features, and the dimension of the features is [m, n]. And $\textit{Detector}(f)$ is 0, 1, which 0 stands for normal examples and 1 is adversarial examples. Linear takes an input and applies a linear transformation, multiplying each input by its corresponding weight, summing them up, and then adding a bias. This layer does not have an activation function. The linear layer is commonly used to map input data to the feature space of the next layer.

A fully connected neural network can be used to address regression, prediction, and classification tasks. Typically, as the number of layers increases, the performance of the neural network model improves. However, due to the powerful data processing capabilities of the attention mechanism in the Transformer, we can achieve excellent results using just a single linear layer.

\subsection{Vision Transformer Interpretability}

We attempt to use visualization techniques to explain the influence of adversarial examples for vision transformer. We employ two methods for explainability in Vision Transformers. One is Attention Rollout\cite{b29}, and the other is Gradient Attention Rollout for class-specific explainability\cite{b30}.

Attention rollout is an intuitive way of tracing the propagation of information from the input layer to the embeddings in the higher layers. In all the layers below, we multiply the attention weight matrices recursively.

\begin{equation}
    \label{eq:7}
    \tilde{A}(l_i) = \left\{   
\begin{array}{ll}  
A(l_i)\tilde{A}(l_{i-1}) & \text{if } i > j \\  
A(l_i) & \text{if } i = j  
\end{array}   
\right.
\end{equation}
In this equation ~(\ref{eq:7}), $\tilde{A}$ is attention rollout, A is B in ~(\ref{eq:4}),and we set $j = 0$.

On the basis of equation~(\ref{eq:4}), the calculation process for the gradient attention rollout method is as follows.

\begin{equation}
    \label{eq:8}
  \hat{A}^i = I + \mathbb{E}_h(\nabla A^i \odot R^{n_i})
\end{equation}

\begin{equation}
    \label{eq:9}
    \textit{Rollout}= \hat{A}^1\cdot\hat{A}^2\dots\cdot\hat{A}^I 
\end{equation}
Where $\mathbb{E}_h$ is the average of multi-head output, $\nabla A^i$ is the gradient of $A^i$ that is attention matrix, and $R^{n_i}$ is the correlation coefficient. The result of the rollout method is fixed and is not affected by changes in the target category.

\subsection{Training \& Loss}
 
In this section, we introduce the training process of our detector and the loss function. When training our classifier, we use the SGDM optimizer to update the model parameters. The input feature is denoted as $X$, the corresponding true labels as $Y$, and the model parameters as $\theta$.  

\begin{equation}
    \label{eq:10}
    \hat{Y} = f(X; \theta)
\end{equation}    
Here, $f$ represents the linear network's output function, which can be expressed as $f(X; \theta) = X \cdot \theta$, where $X$ is the input data matrix, and $\theta$ is the column vector of model parameters. We use Cross Entropy Loss to update parameters, and it is denoted as $L$ :  

\begin{equation}
    \label{eq:11}
    \textit{L}_{CE} = -\frac{1}{N} \sum_{i=1}^{N} \left[ y_i \cdot \log(\hat{y}_i) + (1 - y_i) \cdot \log(1 - \hat{y}_i) \right]
\end{equation}  

\begin{equation}
    \label{eq:12}
    v_t = \beta \cdot v_{t-1} + (1 - \beta) \cdot \nabla_{\theta} \textit{L}
\end{equation}  

\begin{equation}
    \label{eq:13}
    \theta = \theta - \alpha \cdot v_t
\end{equation}  
In equation ~(\ref{eq:11}), $N$ represents the number of examples, $y_i$ is the true label of the $i$-th sample, and $\hat{y}_i$ is the model's predicted output for the $i$-th sample. Following this, we compute the gradient $\nabla_{\theta} L$ of the loss $L$ with respect to the parameters $\theta$. Subsequently, the parameters $\theta$ are updated using SGD with momentum, with equation~(\ref{eq:12}) and (\ref{eq:13}), where $v_t$ represents the velocity (momentum) at time step $t$, $\beta$ is the momentum parameter, and $\alpha$ is the learning rate. This process is repeated until a stopping condition is met, such as reaching the maximum number of iterations or convergence of the loss function.

\begin{table}[htbp]  
\centering  
\caption{Attack parameters}
\resizebox{0.45\textwidth}{!}{%
\begin{tabular}{|c|c|c|c|}  
\hline  
\multicolumn{4}{|c|}{\textbf{ImageNet-val set}} \\  
\hline  
Method & \multicolumn{3}{c|}{\textbf{Parameters}} \\  
\hline  
PGD & \multicolumn{3}{c|}{$\epsilon = 0.0625$,$steps = 10$, $\alpha = 2/255$}\\  
FGSM & \multicolumn{3}{c|}{$\epsilon = 0.0625$}\\  
BIM & \multicolumn{3}{c|}{$\epsilon = 0.0625$,$steps = 10$, $\alpha = 2/255$} \\  
CW & \multicolumn{3}{c|}{$steps = 50$, $c = 1$, $lr = 0.01$, $\kappa = 0$ }\\ 
MIM & \multicolumn{3}{c|}{$\epsilon = 8 / 255$,$steps = 10$, $\gamma = 1.0$} \\ 
Patch-Pool & \multicolumn{3}{c|}{$num_{patch} = 1$, $atten_{select} = 4$, $attack_{mode} = loss_{CE}$, $lr = 0.22$} \\  

\hline  
\end{tabular}  
}
  
\label{tab_parameters}  
\end{table}

\begin{table}[htbp]  
\centering  
\caption{Classification accuracy of three pre-trained ViT models under various adversarial attacks}
\resizebox{0.45\textwidth}{!}{%
\begin{tabular}{|c|c|c|c|}  
\hline  

Model Name & \textbf{ViT-B-16} & \textbf{ViT-B-32} & \textbf{DeiT-Tiny} \\  
\hline  
\text{No attack} & 79.34\% & 75.77\% & 72.34\% \\  
\hline
\text{PGD} & 4.10\% & 5.06\% & 0.86\% \\  
\hline
\text{FGSM} & 39.82\% & 32.69\% & 25.90\% \\  
\hline
\text{BIM} & 18.24\% & 20.55\% & 9.34\% \\  
\hline
\text{CW} & 50.81\% & 42.37\% & 7.98\% \\
\hline
\text{MIM} & 7.17\% & 8.30\% & 2.68\% \\  
\hline 
\text{Patch-fool} & 24.61\% & 36.89\% & 5.11\% \\  
\hline 
\end{tabular}  
}  
  
\label{tab_acc}  
\end{table}

\begin{figure*}[htbp]
\centerline{\includegraphics[width=0.7\textwidth]{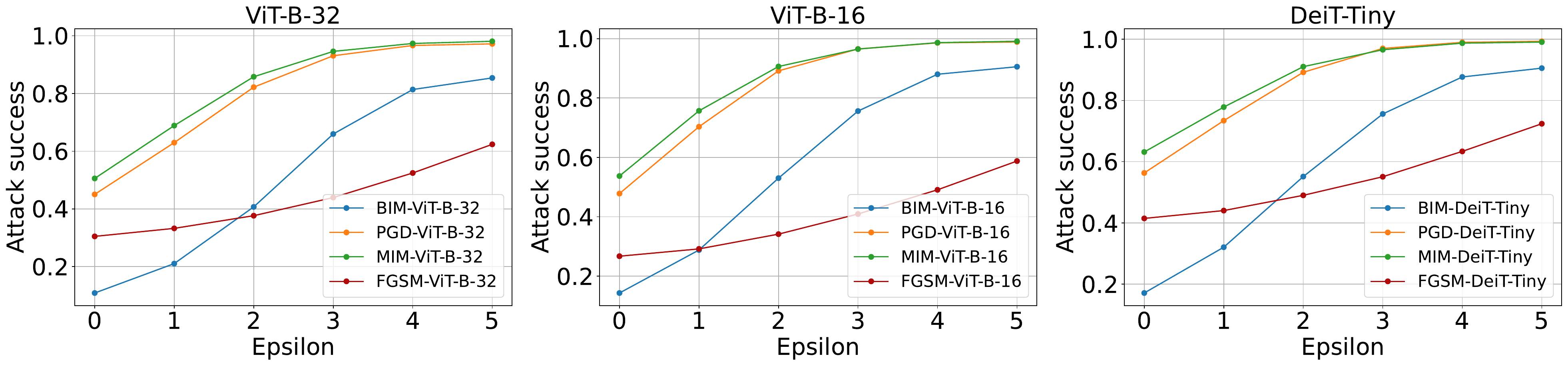}}
\caption{Attack effectiveness for three ViT models. $\epsilon$ are $1/256, 2/256, 4/256, 8/256, 16/256, 32/256$. }
\label{fig_cam}
\end{figure*}

\begin{figure*}[htbp]
\centerline{\includegraphics[width=0.6\textwidth]{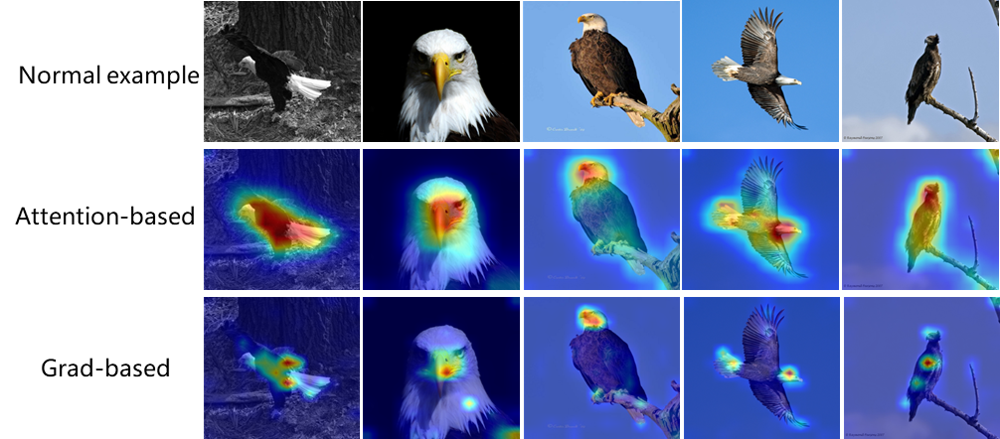}}
\caption{The results obtained using the attention rollout method and the grad attention rollout method on normal examples.}
\label{fig_cam}
\end{figure*}

\begin{figure*}[htbp]
\centerline{\includegraphics[width=0.7\textwidth]{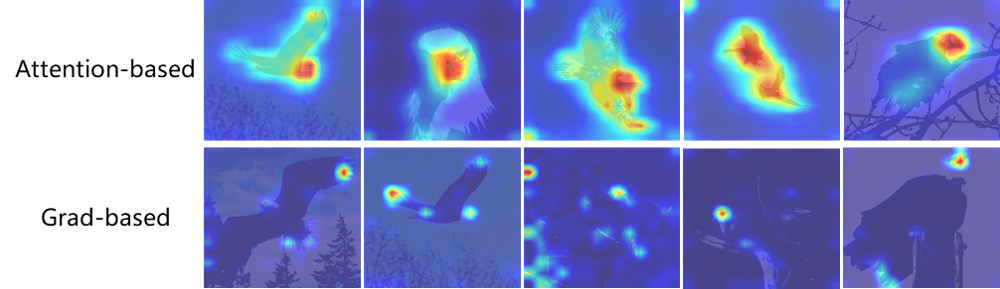}}
\caption{The results obtained using the attention rollout method and the grad attention rollout method on adversarial examples.}
\label{fig_pgdcam}
\end{figure*}

\begin{table*}[htbp]
\centering
\caption{The table evaluates the attack capabilities of different attack methods (such as PGD, FGSM, BIM, etc.) on different models (vit-b-16, vit-b-32, DeiT-Tiny) using three metrics to represent their effectiveness. ASR represents the rate of successful adversarial attacks, while $L_2$ and $L_{\infty}$ norms measure the differences between adversarial examples and normal examples.}

\resizebox{0.9\textwidth}{!}{%
\begin{tabular}{|c|c|c|c|c|c|c|c|c|c|}
\hline
\multirow{2}{*}{\textbf{Attack Type}} & \multicolumn{3}{c|}{\textbf{ViT-B-16}} &  \multicolumn{3}{c|}{\textbf{ViT-B-32}} &  \multicolumn{3}{c|}{\textbf{DeiT-Tiny}} \\ 
\cline{2-10}
  & ASR & $L_2$ & $L_{\infty}$	& ASR & $L_2$ &$L_{\infty}$ & ASR & $L_2$ & $L_{\infty}$ \\
\hline
PGD & 0.946 & 0.5842 & 2.0624 & 0.9283 & 0.5836 & 2.0615 & 0.9877 & 0.5849 & 2.0212 \\  
\hline
FGSM & 0.4757 & 0.5847 & 2.0623 & 0.538 & 0.5832 & 2.0615 & 0.6298 & 0.5856 & 2.0212 \\  
\hline
BIM & 0.7598 & 0.0866 & 1.5471 & 0.7096 & 0.0866 & 1.5454 & 0.8664 & 0.0855 & 1.4827 \\ 
\hline
CW & 0.3308 & 0.496 & 1.735 & 0.4013 & 0.41 & 1.4185 & 0.886 & 0.5165 & 1.7888 \\  
\hline
MIM & 0.9056 & 0.5839 & 2.0622 & 0.8827 & 0.5823 & 2.0615 & 0.9617 & 0.5847 & 2.0212 \\  
\hline
Patch-fool & 0.7572 & 0.00015 & 0.1046 & 0.6311 & 0.0002 & 0.1100 & 0.9489 & 0.0006 & 0.1147 \\  
\hline 
\end{tabular}
}
\label{tab_asr}

\end{table*}

\begin{table*}[htbp]
\centering
\caption{AUC scores of different detection methods against adversarial attacks on ImageNet-val set}

\resizebox{0.6\textwidth}{!}{%
\begin{tabular}{|c|c|c|c|c|c|c|}
\hline
\multirow{2}{*}{\textbf{Attack Type}} & \multicolumn{2}{c|}{\textbf{ViT-B-16}} &  \multicolumn{2}{c|}{\textbf{ViT-B-32}} &  \multicolumn{2}{c|}{\textbf{DeiT-Tiny}} \\ 
\cline{2-7}
  & LID & Ours& LID & Ours & LID & Ours \\
\hline
\text{PGD} &  0.8314 & 0.9976 & 0.7468 & 0.9671 & 0.8101 & 1 \\  
\hline
\text{FGSM} & 0.9011 & 1 & 0.8902 & 0.9971 & 0.8902 & 1 \\ 
\hline
\text{BIM} & 0.8802 & 0.9895 & 0.7012 & 0.9245 & 0.8793 & 0.9997 \\ 
\hline
\text{CW} & 0.6537 & 0.9791 & 0.631 & 0.9552 & 0.7146 & 0.9988 \\  
\hline
\text{MIM} & 0.6998 & 0.9912 & 0.6481 & 0.9812  & 0.7023 & 0.9897 \\  
\hline
\text{Patch-fool} & 0.6861 & 0.9942 & 0.6243 & 0.9311 & 0.6798 & 0.9502 \\  
\hline 
\end{tabular}
}
\label{tab_auc}

\end{table*}

\section{Evaluation}
In this section, we first compare our efficient detection
methods with existing detection methods. Then, we investigate the
impact of different attack methods in three pre-trained ViT models, visualization strategies and evaluation metrics in our methods. Finally, we demonstrate the \sys performs better than other baselines.  

\subsection{Experimental Setting}
\paragraph{Dataset} In this section, we evaluate the performace of \sys on ImageNet dataset\cite{imagenet}, which is one of the most commonly used datasets in the field of deep learning for tasks such as image classification, detection, and localization. Because ViT models are generally used for large-scale datasets, we are only focusing on one dataset in this paper.

Typically, we use the ILSVRC2012 subset of ImageNet, which consists of 1000 categories, with approximately 1000 images per category. The total number of training images is around 1.2 million. Additionally, there are some images allocated for validation and testing. ILSVRC2012 contains 50,000 images for validation and 100,000 images for testing. We process the images into $224 \times 224$ for DeiT-Tiny model, $384 \times 384$ for ViT-B-16 and ViT-B-32 models.

\paragraph{Implementation} The adversarial attack parameters of various attack types is presented in Table~\ref{tab_parameters}. For the first five types of attacks, the meanings of the parameters are as follows. $\epsilon$ is the allowed perturbation size, usually in pixel values or normalized to the $[0, 1]$ range. “steps” represents the number of iterations, i.e., the number of times gradient updates are applied. $\alpha$ denotes the step size in the attack, indicating the magnitude of perturbation to the input at each iteration. $c$ is the quadratic term coefficient in the CW (Carlini and Wagner) loss function\cite{b31}. $\kappa$ is a parameter that controls the balance between adversarial loss and natural loss. $lr$ is the learning rate for gradient descent. The parameters of patch-fool attack is different from previous attacks. $num_patch$ is the number of perturbed patches, attack loss function is CE loss, $atten_{select}$ stands for select patches based on which attention layer.

We adopt three typical pre-trained models to demonstrate the vulnerability of DNN models with various adversarial attacks. In "ViT-B-16/32", B presents base-scale parameter and model trained on patch size 16/32. Another model is popular ViT model, DeiT-Tiny. We select this model represents the variant of the ViT models.

\paragraph{Baseline} \textbf{LID}: When calculating the distance distribution, the number of neighbors is set to 10. 

\paragraph{Metrics} We select attack success rate(ASR) as the metric of measuring the vulnerability of naive models. The pre-trained model is more vulnerable as the ASR increases. The metric AUC used to evaluate performance of the detector. $L_2$ and $L_{\infty}$ norms measure the differences between adversarial examples and normal examples. The calculation method for these metrics is as follows.

\begin{equation}
    \label{eq:14}
    \textit{ASR} = \frac{number_{attack success}}{number_{all}}
\end{equation}  
Where $num_{attack success}$ is the number of attack success examples, and $number_{all}$ is all examples.
\begin{equation}
    \label{eq:15}
    \text{AUC} = \frac{\sum_{i=1}^{n}\text{rank}_{\text{pos}}(i) - \frac{n_{\text{pos}}(n_{\text{pos}}+1)}{2}}{n_{\text{pos}} \times n_{\text{neg}}}
\end{equation}  
In this equation ~(\ref{eq:15}), $n$ represents the total number of examples, $n_{\text{pos}}$ stands for the number of positive examples, $n_{\text{neg}}$ denotes the number of negative examples, and $\text{rank}_{\text{pos}}(i)$ indicates the position of the $i$-th positive sample in the sorted set of examples.

\subsection{Robustness of pre-trained ViT models}

Table ~\ref{tab_acc} shows the classification accuracy of three pre-trained ViT models under different attacks. The second row is benign accuracy of pre-trained models. We conducted attacks on these three models using the validation set of the ImageNet dataset. In the absence of any attack, the ViT-B-16 model demonstrates the highest accuracy, outperforming both the ViT-B-32 and DeiT-Tiny models. However, when exposed to adversarial attacks such as PGD, FGSM, BIM, CW, MIM, and Patch-fool, the DeiT-Tiny model consistently exhibits the lowest accuracy, indicating its heightened vulnerability to such attacks compared to the other two models. Notably, the PGD and MIM attacks notably impact the classification accuracy of all three models, signifying its potency in compromising the models’ robustness. 

Table ~\ref{tab_asr} measures the success of these attacks using metrics such as the attack success rate (ASR), the $L_2$ norm, and the $L_{\infty}$ norm. As with Table ~\ref{tab_acc}, the DeiT-Tiny model consistently shows the highest susceptibility to adversarial attacks across all attack types. This is indicated by its elevated ASR and norm values. This result underscores the varying degrees of susceptibility among the models and provides crucial insights into their robustness under adversarial conditions.

\subsection{Performance of Plugin-Detector}

Table ~\ref{tab_auc} present the performance of our plugin-detector in comparison to feature squeezing(FS) and local intrinsic dimensionality(LID). We compare the AUC of our detector with other baselines. It obvious that our plugin-detector can achieve superior detection performance  against attack types such as PGD, FGSM, BIM, CW, MIM, and Patch-fool. Notably, the plugin-detector host in three pre-trained ViT models achieves perfect discrimination(AUC score is 1) for the Our detection method in some evaluated scenarios.

Both normal and adversarial examples were analysed for feature visualisation. It was observed that the attention-based rollout method covers a larger area compared to the gradient-based rollout method, and is therefore more adept at identifying the features of the sample. Furthermore, there is a strong correlation between classification accuracy and this specific area, as both rollout methods focus on the eagle's head and beak. When the gradation-based rollout method is applied to adversarial examples, it is found that the model emphasises the edges of the images rather than the important areas observed in normal examples, indicating a change in this characteristic due to adversarial examples.

\section{Discussion \& Future Work}

In this section, we discuss the limitations and future work of \sys.
\paragraph{Feature Engineering} The feature processing technique of simply subtracting the features of adversarial examples from those of normal examples is used in this paper. In many cases, this method is not applicable. Feature engineering is the process of transforming raw data into a better representation of the underlying problem, enabling the application of these features to predictive models to improve the model’s predictive accuracy on unseen data. 

\paragraph{Transferability}Limitations of our method include lacking high transferability between different models and lacking experimentation across multiple datasets to verify transferability between different datasets and pre-trained models. This means that we have not yet succeeded in achieving a universal detector.

The cross-modal application is an important development direction for ViT. In the future, we can foresee more research focusing on delving into the techniques and methods of cross-modal application to expand the application scope of ViT and improve its performance, which will face new security challenges. The cornerstone of the metaverse lies in the application of multimodal models. We will explore the security issues of larger-scale multimodal models(MLLM) to ensure the safe development of the metaverse.

\section{Conclusion}

In this paper, we introduce \sys, a framework of adversarial examples detection, aimed at preventing pre-trained ViT model from being manipulated. We leverage the intrinsic capabilities of ViT models to obtain the key feature. Our method is also referred to as “making the model think twice before inferencing.” Our experiments  have validated the efficacy of \sys in three different pre-trained ViT models.

\end{document}